\documentclass[conference]{IEEEtran}
\IEEEoverridecommandlockouts
\usepackage{cite}
\usepackage{amsmath,amssymb,amsfonts}
\usepackage[linesnumbered,ruled,vlined]{algorithm2e}
\usepackage{algorithmic}
\usepackage{graphicx}
\usepackage{textcomp}
\usepackage{balance}
\usepackage{xcolor}
\usepackage{tikz}
\usepackage{multirow}
\usepackage{placeins}
\usepackage{booktabs}
\usepackage{longtable}
\usepackage{amsmath}
\def\BibTeX{{\rm B\kern-.05em{\sc i\kern-.025em b}\kern-.08em
    T\kern-.1667em\lower.7ex\hbox{E}\kern-.125emX}}
\begin{document}

\title{Zero-shot Learning with Minimum Instruction to Extract Social Determinants and Family History from Clinical Notes using GPT Model\\
}

\author{\IEEEauthorblockN{Neel Jitesh Bhate}
\IEEEauthorblockA{\textit{Department of CIS} \\
\textit{IUPUI}\\
Indianapolis, IN, USA \\
nbhate@iu.edu}
\and
\IEEEauthorblockN{Ansh Mittal}
\IEEEauthorblockA{\textit{Department of CIS} \\
\textit{IUPUI}\\
Indianapolis, IN, USA \\
anshmitt@iu.edu}
\and
\IEEEauthorblockN{Zhe He}
\IEEEauthorblockA{\textit{School of Information} \\
\textit{Florida State University}\\
Tallahassee, FL, USA \\
Zhe.He@cci.fsu.edu}
\and
\IEEEauthorblockN{Xiao Luo}
\IEEEauthorblockA{\textit{Department of MSIS, Department of CIT} \\
\textit{Oklahoma State University, IUPUI}\\
Stillwater, OK,\ Indianapolis, IN, USA \\
xiao.luo@okstate.edu, luo25@iu.edu}
}

\maketitle

\begin{abstract}
Demographics, Social determinants of health, and family history documented in the unstructured text within the electronic health records
 are increasingly being studied to understand how this information can be utilized with the structured data to improve healthcare outcomes. After the GPT models were released, many studies have applied GPT models to 
 extract this information from the narrative clinical notes. Different from the existing work, our research focuses on investigating the zero-shot learning on extracting this information together by providing minimum information to the GPT model. 
We utilize de-identified real-world clinical notes annotated for demographics, various social determinants, and family history information. Given that the GPT model might provide text different from the text in the original data, 
we explore two sets of evaluation metrics, including the traditional NER evaluation metrics and semantic similarity evaluation metrics, to completely understand the performance. Our results show that the GPT-3.5 method achieved an
average of 0.975 F1 on demographics extraction, 0.615 F1 on social determinants extraction, and 0.722 F1 on family history extraction. We believe these results can be further improved through model fine-tuning or few-shots learning. Through the case studies, we also identified the limitations of the GPT models, which need to be addressed in future research.
\end{abstract}

\begin{IEEEkeywords}
GPT, Generative AI, clinical text, social determinants, family history
\end{IEEEkeywords}

\section{Introduction}

Although many data elements in the electronic health records can be stored in structured fields, such as date of birth, gender, race, ethnicity, diagnosis, medication, etc., some elements that are strongly related to patients' health conditions are not fully captured by the structured fields, including social determinants and family history. Social determinants are the words describing the social history of health, such as employment status, smoking history, and alcoholism, etc. Family history is the description of the disease history of the direct or indirect family members. Also, some elements, such as ethnicity, are not recorded in the structured data fields. Because of the long history of paper-based charting, physicians often write important clinical information in the narrative clinical notes. There is a critical need to extract social determinants and family history from the narratives to complement other clinical features for disease risk stratification and management. 

With recent advances in natural language processing, other than the rule-based models \cite{almeida2020rule,patra2021extracting}, various transformer-based language models \cite{patra2021extracting,romanowski2023extracting,yang2020extracting} have been developed to extract the social determinants and family history from the narrative clinical notes. However, one limitation of the existing models is that they require extensive human labor to label the narrative clinical text to prepare the training data. Recent developments in Large Language Models (LLMs) like the GPT-3.5, GPT-4, med-PaLM, and Flan-PaLM models have shown their capability to understand the clinical text and some exceed human-level performance in different tasks including clinical name entity recognition \cite{ramachandran2023prompt,hu2023zero,nori2023capabilities}, clinical question answering \cite{nori2023capabilities}, and clinical image to text generation \cite{selivanov2023medical}, etc.  However, it is still an open question whether zero-shot learning using GPT models can extract social determinants and family history with reasonable performance. Given that social determinants and family history are different from the other clinical terminologies which require extensive and dedicated vocabulary as the backend support, we hypothesize that the GPT models might gain comparable performances with zero-shot learning. 

In this work, we explore the extraction of social determinants, family history, and demographic information using zero-shot learning based on GPT-3.5. Unlike the existing approach using the GPT model for entity extraction, we do not use prompts to generate the output as inline annotations. Instead, we provide prompts to allow GPT model to output phrases with different terms than the inline annotations but with high semantic similarity. Since there is no existing standard for representing social determinants and family history in the EHR clinical notes, and one term can be represented in various ways, zero-shot learning without detailed instruction might benefit the task without a large amount of labeled data.  
Although we used a different dataset than some study in the literature \cite{romanowski2023extracting,torii2023task,torii2023task}, our experiments show that GPT model has a big potential to extract social determinants and family history information from clincial text without any training data.

\begin{table*}[h!]\label{annotation1}
    \centering
     \caption{Annotation examples of age, gender, ethnicity, and family history}
    \begin{tabular}{c|c|c|c|c|c}\hline
        \multicolumn{6}{l}{Sentence 1: Family history: father had heart disease and died} \\
        \multicolumn{6}{l}{Sentence 2: 76 year old white woman} \\
        \multicolumn{6}{l}{Sentence 3: 54-year-old Chinese female without significant past medical history} \\
        \hline
        \multirow{2}{*}{} & \multirow{2}{*}{Age} & \multirow{2}{*}{Gender} & \multirow{2}{*}{Ethnicity} & \multicolumn{2}{|c}{Family History}\\\cline{5-6}
          & & & & Observation & Vital\\\hline
          Sentence 1 & & & & father had heart disease & died \\\hline
          Sentence 2 & 76 year old & female & White & & \\\hline
          Sentence 3 &  54 year old & female & Chinese & & \\\hline
    \end{tabular}

\end{table*}

\begin{table*}[h!]
    \centering
    \caption{Annotation examples of social history}
    \label{annotation2}
    \resizebox{\textwidth}{!}{\begin{tabular}{c|c|c|c|c|c}
    \hline
    \multicolumn{6}{l}{Sentence 1: Non-smoker, never, denies drug use, She is married, has 3 children 2 grandchildren. has worked in the school system.} \\
    
    \multicolumn{6}{l}{Sentence 2: Unemployed, Some college education, Yoga exercise, Lives with daughter at Salvation Army post, Separated with 2 children, Never smoker, } \\
   
     \multicolumn{6}{l}{\ \ \ \ \ \ \ \ \ \ \ \ \ \ \ Denies alcohol and drug use} \\
    \multicolumn{6}{l}{Sentence 3: Lives with spouse and 2 children, employed as a fiscal coordinator, never smoker } \\
   \hline
    \multicolumn{6}{c}{Social History}\\\hline
    Employment status & Alcohol use & Tobacco use & Drug use & Education status & Living status \\\hline
    worked in the school system &  &  non-smoker& denies drug use &  &  \\\hline
    unemployed & denies alcohol and drug use &  & denies alcohol and drug use & some college education & lives with daughter \\\hline
    employed as a fiscal coordinator &  & never smoker &  &  & lives with spouse and 2 children \\\hline
    
    \end{tabular}}
\end{table*}

\begin{table}[h!]\label{sum}
    \centering
     \caption{Statistics of the Data}
    \begin{tabular}{c|c|c}\hline
        
Type & sub-type & Count  \\\hline
Age & - & 680  \\\hline
Gender & - & 840 \\\hline
Ethnicity & - & 69 \\\hline
Social History & Employment status & 187 \\
 & Alcohol use & 368 \\
 & Tobacco use & 756 \\
 & Drug use & 61 \\
 & Education status & 40 \\
 & Living status & 291 \\\hline
 Family History & Observation & 612 \\
   & Vital & 118 \\

 \hline
    \end{tabular}
   
\end{table}

\section{Data and Annotation}
To evaluate the GPT model, we extracted 1,000 narrative clinical notes from more than 150 patients with different diseases from an University hospital. We focus on primary care MD progress notes. For this research, we have de-identified the clinical notes by removing all the personal information, including mentioned names, date of birth, addresses, phone numbers, and medical record numbers etc. Then, two independent annotators labeled the social determinants based on the guideline summary for all event types listed in \cite{lybarger2021annotating}, and family history based on the guideline given in \cite{bill2014automated}. A third annotator worked on solving the conflicting cases between the two main annotators.  Table I shows the annotation examples of age, gender, ethnicity, and family history. Table II shows the annotation examples of the social history. Table III summarizes the number of events for each type.

\section{Method}
\subsection{Information Extraction using GPT model}
We conduct the GPT-3.5 zero-shot experiments through OpenAI GPT-3.5 Application Programming Interface (API) \footnote{https://platform.openai.com/docs/api-reference/}. The API allows users to provide instructions via two role variables. Our prompt is intuitively structured in the following order:
\begin{itemize}
    \item  System: defines task instructions for GPT in the desired role. We use
the system variable to provide task instructors so that the model acts as the role of
an annotator.
\item User: provides an input text for zero-shot learning
\end{itemize}
Following the above definitions, we end with a user message containing a clinical note to be annotated and indicating that the assistant should respond. Figure 1 shows the design of the prompt used in this research. The prompt is simple with minimum information to guide the GPT model to process the input text and output the results in a given format which includes the main types to be extracted. The reason to define such a simple and intuitive prompt is to investigate the performance of the GPT-3.5 model on clinical text processing with minimum instruction.  

\begin{figure}[h!] \label{prompt}
  \centering
  \includegraphics[width=0.5\textwidth]{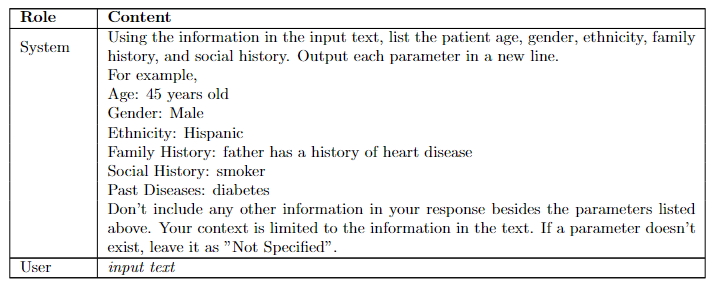}
  \caption{Designed prompt for the study}
\end{figure}

\begin{table*}[h!]
    \centering
    \caption{Performance Evaluation Results on Each Entity Type}
    \label{annotation2}
    \begin{tabular}{c|c|c|c|c|c|c}
    \hline

     & \multicolumn{3}{|c}{NER evaluation metrics} & \multicolumn{3}{|c}{Semantic evaluation metrics}\\\hline
     & P & R & F1 & Ave ($s$) & Acc ($\theta=0.8$) & Acc ($\theta=0.9$)\\\hline
    Age & 0.991 & 0.992 & 0.991 & 0.996 & 0.993 & 0.991 \\
    Gender & 0.977 & 0.977 & 0.977 & 0.990 & 0.986 & 0.983\\
    Ethnicity & 0.948 & 0.949 & 0.948 & 0.938 & 0.913 & 0.909\\
    Social History & &&&&& \\
    Employment status & 0.632 & 0.574 & 0.613 & 0.788 & 0.652 & 0.512\\
    Alcohol use & 0.683 & 0.614 & 0.652 & 0.831 & 0.724 & 0.594\\
    Tobacco use & 0.662 & 0.631 & 0.652 & 0.848 & 0.747 & 0.649\\
    Drug use & 0.609 & 0.579 & 0.603 & 0.829 & 0.742 & 0.583\\
    Education status & 0.583 & 0.551 & 0.562 & 0.779 & 0.677 & 0.378\\ 
    Living status & 0.636 & 0.568 & 0.608 & 0.793 & 0.642 & 0.504 \\
    Family History &  &&&&& \\
    Observation & 0.708 & 0.702 & 0.704 & 0.713 & 0.515 & 0.478\\
    Vital & 0.739 & 0.739 & 0.739 & 0.739 & 0.529 & 0.452\\\hline
    \end{tabular}
\end{table*}

\subsection{Post-processing}
Since our objective is to provide minimum information to the GPT model, the instruction about the subtypes of the social history and family history are not given. In order to calculate the performance on those subtypes, we employ a postprocesing step. If a string is extracted as one type, e.g., social history, it is then further broken down into segments based on commas and other punctuation. For each segment, the presence of subtype-specific keywords are used to categorize them into subtypes. For example, if a string ``unemployed, without smoking history, some college education'' is extracted as social history, it is broken into segments ``unemployed'', ``without smoking history'' and ``some college education''. To determine if a segment is relevant to ``education status'', keywords such as `school', `college', `university', `graduate', `degree', etc., are used.

\section{Evaluation Metrics}
Although the existing research \cite{hu2023zero} compares the LLMs on clinical information extraction using the traditional evaluation metrics that are applied to the Named Entity Recognition (NER) models, we think for the zero-shot learning, it is also informative if semantic similarity-based evaluation is used for comparison. For example, if the annotated ground truth in the original text is ``has no job'', the extracted concept is ``unemployed'', we consider the extraction is semantically correct.  Hence in this study, we employ both NER evaluation metrics and a semantic evaluation approach. 

For the NER evaluation metrics, we use the relaxed match. It is similar to the fragment match in the related research on biomedical named entity recognition \cite{seki2003probabilistic}\cite{tsai2006various}. For the relaxed match, all extracted concepts are converted into words. Then, we calculate the word-based precision and recall. The words that are in both the annotated ground truth and the GPT output would be considered as True Positives (TP). The words that are in the GPT output but not in the annotated ground truth are considered as False Positives (FP), and the words that are in the annotated ground truth but not in the GPT output are considered as False Negatives (FN).  The Equations \ref{eq1} to \ref{eq3} show how the recall ($R$), precision ($P$), and F1 are calculated. 

\begin{equation}\label{eq1}
    P= \frac{TP}{TP+FP}
\end{equation}
\begin{equation}\label{eq2}
  R= \frac{TP}{TP+FN}  
\end{equation}
\begin{equation}\label{eq3}
    F1= \frac{2PR}{P+R}
\end{equation} 

Given the annotated ground truth ``Employed part-time as an accountant, lives with spouse, follows a diet for IBS, former smoker (quit greater than 1 year ago), does not engage in substance abuse, drinks alcohol, exercises 3-4 times a week by walking'', if only ``former smoker (quit greater than 1 year ago)'' is extracted by the GPT model, recall is 1, and precision is 1 for the smoking status. But the recall and precision are 0 for the subtypes of alcohol use (``drinks alcohol''), drug use (``does not engage in substance abuse''), employment status (``Employed part-time as an accountant''), and living status (``exercises 3-4 times a week by walking''). 

For the semantic evaluation, we first convert the annotated ground truth and extracted concept into embeddings using the sentence-transformers model -- all-MiniLM-L6-v2 \cite{reimers2019sentence}, then we measure the cosine similarity between the embeddings. If the cosine similarity ($s$) is above a threshold $\theta$ (e.g., $\theta=0.8$), it is a match. Otherwise, it is counted as a wrong extraction. We calculate the average cosine similarity (Ave ($s$)) for each type without adding a threshold and calculate the accuracy using the threshold 0.8 (Acc ($\theta=0.8$)) and 0.9 (Acc ($\theta=0.9$)), respectively. When $\theta$ is set to be 0.9, the result is close to NER exact match which means all words in the annotated ground truth and the GPT output are the same.

\section{Experimental Results}
\subsection{Performance Evaluation}
Table IV shows the performances of each type and subtype using both NER and semantic evaluation metrics. The results show that the model performs very well on demographic information extraction. The NER evaluation and semantic evaluation both show that the GPT model can extract detailed information correctly. Whereas the performance on social determinants and family history extraction is lower. The NER F1 measure shows that the performance on most of the subtypes of social determinants is around 0.6, and the performance on the two subtypes of the family history is a little above 0.7. However, it is worth noting that these results demonstrate the GPT model can identify and extract the information relevant to social determinants and family history with very minimum instruction. If semantic evaluation with threshold $\theta=$0.9 is used, it is the result of an exact match. The performance on education status is the lowest, with accuracy of 0.378, and tobacco use is the highest, with an accuracy of 0.649. 

\subsection{Case Study}
We use three scenarios that mirror the degrees of alignment between the annotated ground truth and the output generated by GPT.  Case 1, shown in Figure 2, is an example of family history. The alignment between annotated ground truth and GPT output is high. However, the GPT did not include the ``Cancer (aunt, maternal grandmother, paternal grandmother)''. However, when we input this part to process again, GPT keeps it part of the family history. This means the response from GPT is not stable. It can change from time to time. 

\begin{figure}[h!] \label{case1}
  \centering
  \includegraphics[width=0.45\textwidth]{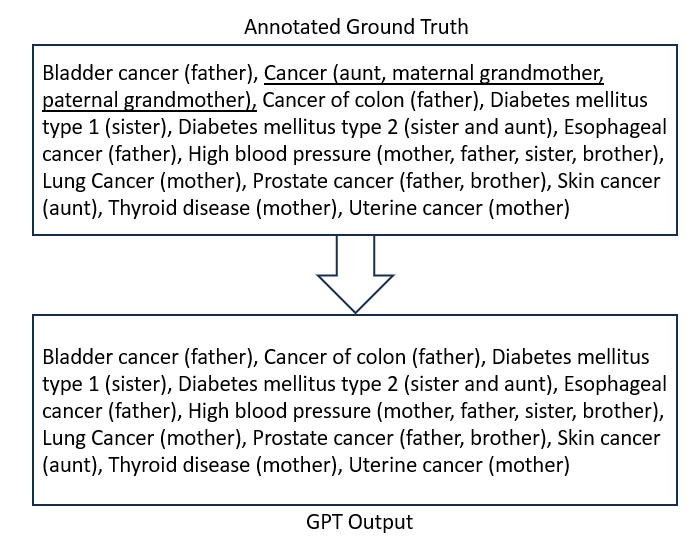}
  \caption{Case 1 -- Family History}
\end{figure}

Case 2, shown in Figure 3, is an example of social history. The difference between the annotated ground truth and the output of GPT is small. However, we noticed that the semantic meaning of the output of GPT is the same as the annotated ground truth. But GPT added a few words to make the content more complete. This demonstrates that GPT may not keep the original content but may generate new content with the same semantic meaning. 

\begin{figure}[h!] \label{case1}
  \centering
  \includegraphics[width=0.43\textwidth]{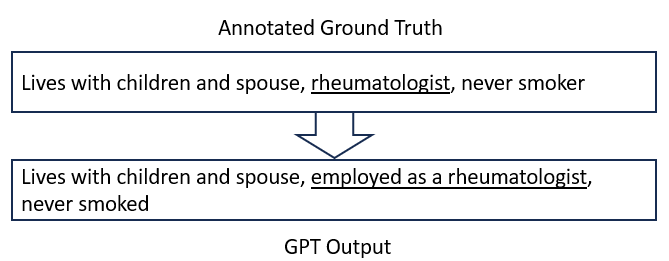}
  \caption{case 2 -- Social Determinant}
\end{figure}

Case 3, shown as Figure 4, is another example of social history. GPT missed a few social history elements. After checking the original content, we found that the original content was spread out in different section in the clinical notes, and some of the content are tagged using a category label with ``:'', but the style is not consistent within the text. This might make it challenging for GPT to understand the content. 

\begin{figure}[h!] \label{case1}
  \centering
  \includegraphics[width=0.43\textwidth]{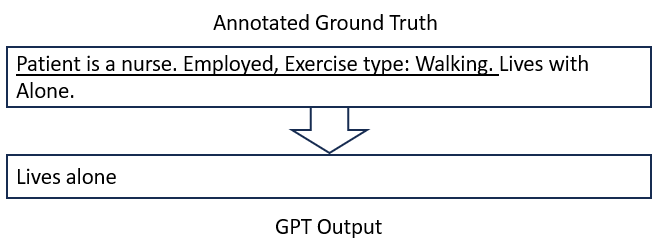}
  \caption{Case 3 -- Social Determinant}
\end{figure}

\subsection{Comparison with Approaches in Literature}
Although we used our own data set, it is important to compare our results with some related work in the literature.  We found that we gained similar performance as other studies in the recent literature, although we took a simple approach. Hu et al. \cite{hu2023zero} applied GPT-3.5 in the zero-shot setting to extract medical problems, treatments, and tests from the clinical notes. They obtained F1 scores of 0.418 (vs.0.250) and 0.620 (vs. 0.480) using the NER relaxed and exact matching. However, they designed the prompts to extract without allowing the GPT model rephrasing the content. Since the medical problems, treatments, and tests include many professional and clinical domain terminologies, it is reasonable to design prompts with more instructions. Ramachandran et al. \cite{ramachandran2023prompt} applied few-shot
learning using the GPT-4 models. Although much more specific instructions are given, the overall F1 on social determinants extraction using the GPT-standoff and GPTinline models are 0.625 and 0.652, respectively. Our overall performance on social determinants extraction is competitive (0.611), although we used zero-shot learning with simple instruction. 

\section{Conclusions}
Our research results show that with minimum instruction, the GPT model can achieve high F1 on demographic information extraction  (0.975), and moderately high F1 on social determinants extraction (0.615), and  family history extraction (0.722). This demonstrates that GPT models have a great potential to extract entities from the clinical text. Our case study shows that the GPT models have limitations, which are also shown by other researchers in the recent literature. 

\section{Challenges, Opportunities and Future Impact}
Based on this study, we realize that there are still many challenges in adapting the GPT models or generative AI for clinical natural language processing. First of all, the output of the GPT is not consistent. It is referred to as low self-consistency in the recent research \cite{mitchell2022enhancing,jain2023self}. This means the results are not consistent for the same input. This might be resolved by taking different approaches, such as applying GPT model multiple times and integrating the top-k ranked outputs \cite{jain2023self}, or adding natural language inference \cite{mitchell2022enhancing}. Second, even with detailed instruction, the output of GPT might not be the same as the original content; GPT can generate content based on its trained model \cite{hu2023zero}. Sometimes the GPT model tries to keep the same semantic meaning, sometimes the output differs from the original meaning. This means post-processing, such as validation and normalization, is needed. On the other side, GPT model does not provide source information to the output \cite{goyal2022news,zhang2023complete}, thus making the validation challenging. Third, although our task involves less clinical domain terminology, domain knowledge is often needed to guide the learning of GPT model. 

Even though there are many challenges in applying GPT model for information extraction from clinical text, there are opportunities to better utilize the GPT model by taking the following approaches: (1) Applying an active learning process to involve human in the loop to fine-tune a GPT model. This process enables interaction between experts and the GPT model so that GPT model can be fine-tuned according to the needs of the specific task and domain; (2) Enforcing the GPT output respects the original meaning of the content. Reinforcement learning can be used to fine-tune GPT model to generate results according to the semantic meaning in the input. For example, the semantic similarity between the original content and GPT output can be used to provide rewards for reinforcement learning; (3) Adding post-processing to the output of the GPT model by providing the references to the original input text. For some use cases, it is acceptable for GPT to derive a summary of the original input. However, to validate the output, a post-processing can be added to confirm the source of the output. 

The impact of the GPT models on clinical research is notable. The recent publications have already shown that GPT models can be adapted to various clinical information extraction \cite{agrawal2022large}, clinical question and answering \cite{johnson2023assessing}, and clinical education and recommendations \cite{dash2023evaluation}, etc. With the increasing research in the area of generative AI, there will be increasing impact of the customized GPT models in various clinical information processing and research. This will also bring changes to the traditional workflow of physicians, and change the behaviors of patients, and improve healthcare outcomes.

\section*{Acknowledgment}
This research was supported by the IUPUI MURI program and partially supported by the National Institute of Health (grants 1R15GM139094, 5R21LM013911-02, 5P01AA029547-02).

\bibliographystyle{plain}
\balance
\bibliography{ref}

\end{document}